\documentclass[10pt,twocolumn,letterpaper]{article}

\usepackage{cvpr}              

\usepackage[dvipsnames]{xcolor}

\definecolor{cvprblue}{rgb}{0.21,0.49,0.74}
\usepackage[pagebackref,breaklinks,colorlinks,citecolor=cvprblue]{hyperref}
\usepackage{graphicx}
\usepackage{amssymb}
\usepackage{amsmath}
\usepackage{algorithm}
\usepackage{algpseudocode}
\usepackage{multirow}
\usepackage{array}
\usepackage{wrapfig}
\def\paperID{337} 
\def\confName{CVPR}
\def\confYear{2024}

\title{Distilled Datamodel with Reverse Gradient Matching}

\author{\bf Jingwen Ye \quad Ruonan Yu \quad Songhua Liu \quad Xinchao Wang  $^{\dagger}$ \thanks{ $^{\dagger}$ Corresponding author.}\\
National University of Singapore\\
{\tt\small  jingweny@nus.edu.sg, \{ruonan,songhua.liu\}@u.nus.edu, 
xinchao@nus.edu.sg}
}
\makeatletter
\def\thanks#1{\protected@xdef\@thanks{\@thanks
\protect\footnotetext{#1}}}
\makeatother
\begin{document}
\maketitle
\begin{abstract}

The proliferation of large-scale AI models trained on extensive datasets has revolutionized machine learning. With these models taking on increasingly central roles in various applications, the need to understand their behavior and enhance interpretability has become paramount.
To investigate the impact of changes in training data on a pre-trained model, a common approach is leave-one-out retraining. This entails systematically altering the training dataset by removing specific samples to observe resulting changes within the model. However, retraining the model for each altered dataset presents a significant computational challenge, given the need to perform this operation for every dataset variation.
In this paper, we introduce an efficient framework for assessing data impact, comprising offline training and online evaluation stages. During the offline training phase, we approximate the influence of training data on the target model through a distilled synset, formulated as a reversed gradient matching problem. For online evaluation, we expedite the leave-one-out process using the synset, which is then utilized to compute the attribution matrix based on the evaluation objective.
Experimental evaluations, including training data attribution and assessments of data quality, demonstrate that our proposed method achieves comparable model behavior evaluation while significantly speeding up the process compared to the direct retraining method.
\end{abstract}    
\section{Introduction}
In the contemporary landscape of machine learning and artificial intelligence, our substantial reliance on large-scale training data has become increasingly pronounced. The notable successes of large AI models like GPT-3~\cite{brown2020language}, BERT~\cite{devlin2018bert}, and DALL-E~\cite{ramesh2021zero} can predominantly be attributed to the availability of extensive datasets, enabling them to discern complex patterns and relationships. As AI models progressively embrace a data-driven paradigm, comprehending the notion of ``training data attribution" within a machine learning framework emerges as pivotal. It is imperative to acknowledge that model errors, biases, and the overall capabilities of these systems are frequently intertwined with the characteristics of the training data, making the enhancement of training data quality a reliable avenue for bolstering model performance.

Despite the various techniques available for 
interpreting models' decision-making processes,  
the very most of them concentrate on assessing the significance of features~\cite{ribeiro2016should,lundberg2017unified,selvaraju2017grad} 
and explaining the internal representations of models~\cite{kim2018interpretability,friedman2001greedy,apley2020visualizing,XingyiICCV23}.
When examining the attribution of training data, a persistent dilemma surfaces, one that revolves around the delicate balance between computational demands and effectiveness. On one hand, techniques like influence approximation~\cite{guo2020fastif, koh2017understanding} prioritize computational efficiency, but they may exhibit unreliability, especially in non-convex environments.
Concurrently, another line of research has achieved remarkable progress in approximating the impact of even minor alterations, such as the removal of a single data point or a small subset from the complete training set, on the trained model~\cite{Wu2020DeltaGradRR, pruthi2020estimating}. These methods, however, are tailored specifically for scenarios involving minor changes in the training data, lacking the necessary flexibility for broader applications.

In this study, we prioritize flexibility and robustness by opting to retrain the model using a dataset that excludes specific data points. Subsequently, we compare the newly trained models with the original model. The attribution matrix is then computed based on the specific objectives of model evaluation.
Specifically, to effectively and explicitly study the newly trained models, 
we introduce in this paper
a novel  \emph{Distilled Datamodel} framework (DDM).
The DDM framework is centered around the estimation of parameters for the newly trained models, rather than solely focusing on the evaluation of prediction performance at a specific test point. This approach grants the flexibility to analyze various aspects of model behavior and performance.
As is shown in Fig.~\ref{fig:frame}, DDM encompasses two distinct processes: offline training and online evaluation.
During offline training, we distilled the influence of the training data back to the input space to get a rather small synset, a process achieved through reversed gradient matching. We contend that this novel reversed gradient matching approach, when compared to the standard gradient matching~\cite{zhao2021dataset}, is more effective in afterward mitigating the influence of specific training data on the target network.
During online evaluation, we perturbed the synset by deleting, which, along with the target network, is leveraged to quickly train the new model. With all the newly trained networks, the attribution matrix can be easily obtained for different evaluation objectives.
In a word, our contributions are:
\begin{itemize}
    \item We explore a training data attribution framework that explicitly identifies a training sample's responsible for various behaviors exhibited by the target model. By quantifying the impact and contribution of individual samples, our framework provides  insights into the relationship between the training data and the model.
    \item We introduce a novel influence-based dataset distillation scheme that matches the reversed gradient update,
    which results in a highly efficient unlearning of certain data points from the target network.
    \item Experimental results demonstrate that the proposed analysis method provides an accurate interpretation and achieves a significant speedup compared to its unlearning counterpart.
\end{itemize}
\section{Related Work}

\subsection{Data-based Model Analysis}

Model behavior analysis has emerged as a foundational aspect of machine learning and artificial intelligence research and development, often categorized into training data based and testing data based methods

Testing data based methods focus on elucidating the model's inference capabilities for for a certain input. 
Plenty researches~\cite{altmann2010permutation,Friedman2001GreedyFA,Apley2020VisualizingTE,staniak2018explanations,Ustun2015SupersparseLI,Caruana2015IntelligibleMF,Wei2019GeneralizedLR,sundararajan2017axiomatic,zeiler2014visualizing,Zhou2016LearningDF,Selvaraju2019GradCAMVE,Ribeiro2016WhySI} contribute to this field of research. 

In this study, our primary focus is on analyzing the model's behavior based on its training data, with one key approach being the utilization of influence approximation techniques as demonstrated by prior research~\cite{guo2020fastif,koh2017understanding,Schioppa2021ScalingUI,bae2022if}. As pointed out by the authors,
these approaches primarily focus on local changes that are \emph{infinitesimally-small},
which are also extremely time consuming.
Datamodels~\cite{ilyas2022datamodels} is proposed for analyzing the behavior of a model class in terms of the training data, which measures the correlation
between true model outputs and attribution-derived predictions for those outputs.
Following this work, ModelPred~\cite{zeng2023modelpred} is proposed for predicting the trained model parameters directly instead of the trained model behaviors.
Nevertheless, both these methods still entail the training of a considerable number of models, often in the thousands or tens of thousands, for effectiveness. In this work, we investigate a more efficient framework to facilitate this process.

\subsection{Machine Unlearning}
The concept of unlearning is firstly introduced by Bourtoule \textit{et al.}~\cite{Bourtoule2021MachineU}, which 
aims to eliminate the effect of data point(s) on the already trained model. 
Along this line, machine unlearning has attracted more attentions, of which the existing approaches can be roughly divided into \textit{exact}~\cite{Bourtoule2021MachineU,Cao2015TowardsMS,Ginart2019MakingAF,Brophy2021MachineUF} methods and \textit{approximate} methods~\cite{Brophy2021MachineUF,nguyen2020variational,sekhari2021remember,He2021DeepObliviateAP,Ye2023PartialNC,JingwenECCV22}. 

Exact methods decrease the time it takes to exactly/directly retrain the models.
Bourtoule \textit{et al.}~\cite{Bourtoule2021MachineU} propose an unlearning framework that when data needs to be unlearned, only one of the constituent models whose shards contains the point to be unlearned needs to be retrained.
Cao~\textit{et al.}~\cite{Cao2015TowardsMS} transform learning algorithms used by a system into a summation form and to forget a training data sample, they simply update a small number of summations.
DaRE trees~\cite{Brophy2021MachineUF} are proposed to enable the removal of training data with minimal retraining, which cache statistics at each node and training data at each leaf to update only the necessary subtrees as data is removed.
Unlike the exact methods, the approximate ones try to find a way to approximate the retraining procedure.
To minimize the retraining time, data removal-enabled forests~\cite{Brophy2021MachineUF} are introduced as a variant of random forests, which delete data orders of magnitude faster than retraining from scratch while sacrificing little to no predictive power. 
Nguyen \textit{et al.}~\cite{nguyen2020variational} study the problem of approximately unlearning a Bayesian model from a small subset of the training data to be erased.

The above unlearning methods focus more on balancing the accuracies and the efficiency. Here we focus more on the efficiency, which model the network behavior for analyzing the attributions of a target model.

\begin{figure*}[htp]
  \centering
  \includegraphics[width=1\textwidth]{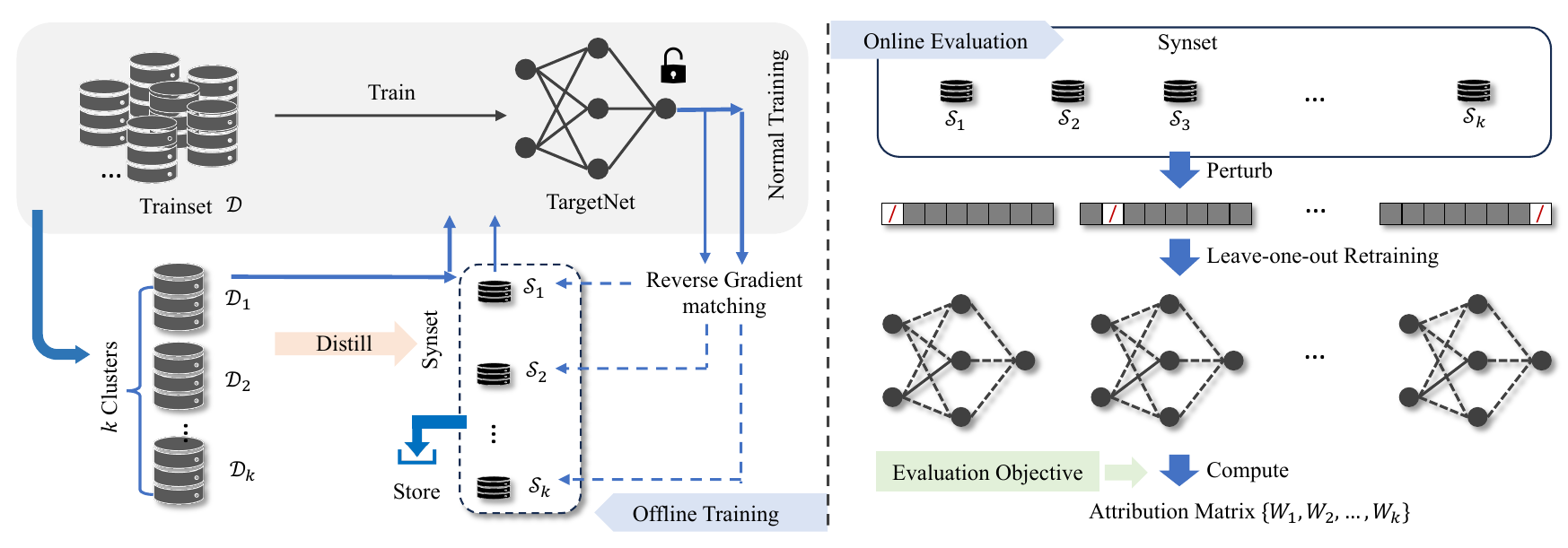}
  \caption{The framework of the proposed distilled datamodel. During the offline training, the synset is distilled during the normal training of target network. As for online evaluation we perturb the learned synset and fast learn the perturbed model set, which is computed to form the final attribution matrix.}
  \label{fig:frame}
\end{figure*}

\subsection{Dataset Distillation}

Dataset condensation/distillation \cite{zhao2021DSA,lee2022dataset,SonghuaNeurIPS22,liu2023slimmable,SonghuaNeurIPS23,RuoNanTPAMI24} aims to condense a large training set into a small synthetic set to obtain the highest generalization performance with a model trained on such small set of synthetic images. Zhao et al. \cite{zhao2021DSA} formulate the goal as a gradient matching problem between the
gradients of deep neural network weights that are trained on the original and the
synthetic data. Zhou et al.~\cite{zhou2022dataset} address these challenges of significant computation and memory costs by neural feature regression with pooling. 
Nguyen et al. \cite{nguyen2021dataset}
apply a distributed kernel-based meta-learning framework to achieve state-of-the-art results for dataset distillation using infinitely wide convolutional neural networks. Sucholutsky et al. \cite{sucholutsky2021soft} propose to simultaneously distill both images and their labels, thus assigning each synthetic sample a ‘soft’ label.

Different from the previous data condensation methods, we tend to use the fast convergence and the gradient matching properties for the analysis of the target network.
Thus, we focus on how to model the data's impact on the network not just for improving the accuracies.

\section{Proposed Method}

In this paper, we propose the distilled datamodel framework to build the training data attribution to evaluate various model behaviors.

\subsection{Problem Statement}
Given a target model $\mathcal{M}$ trained on dataset $\mathcal{D}$, we tend to construct direct relationship between them, which is denoted as the attribution matrix $W$. Each weight in $W$ measures the responsible of the corresponding training points on certain behaviors of $\mathcal{M}$.

The attribution matrix $W$ learned by the proposed DDM framework works on  various behaviors, which include but not limited to:
\begin{itemize}
    \item \textbf{Model functionality analysis.} This involves evaluating the performance of the target network using the training data. This could include measuring key metrics such as accuracy, precision, recall, and F1 score, and comparing the results to established benchmarks or industry standards.
    \item \textbf{Model diagnose.} This involves examining the errors made by the target network when processing the training data. This could include identifying the types of errors made, such as misclassifications or false positives, and determining the root cause of the errors, such as data quality issues or model limitations.
    \item \textbf{Influence function of certain test samples.}  This traces a model's prediction through the learning algorithm and back to its training data, thereby identifying training points most responsible for a given prediction.
\end{itemize}

In what follows, we take studying the model behavior on the influence of certain test samples as an example, showing how to learn the corresponding training data attribution with the proposed DDM framework. 
We would also include more details on  studying other kinds of model behaviors in the supplementary.

Note that in Fig.~\ref{fig:frame}, the proposed DDM framework introduces a two-step process:
\begin{itemize}
    \item \textbf{Offline Training} (Sec. \ref{sec:offline}): This step is learned only once and can be integrated into network training. Its objective is to distill and store data influence with improved approximation and reduced storage requirements. 
    \item \textbf{Online Evaluation} (Sec. \ref{sec:online}): This phase involves evaluation to meet specific requirements for model behavior analysis, which is realized by perturbing the dataset. The primary goal is to compute the training data attribution matrix while minimizing time and computational costs.
\end{itemize}

\subsection{Offline Training}
\label{sec:offline}
During the offline training, we tend to obtain the synset $\mathcal{S}$ ($|\mathcal{S}|\ll |\mathcal{D}|$) to distill the training data influence from the target network $\mathcal{M}$, so as to produce the parameters of the network with perturbed dataset. 
To begin with, we cluster the original training data $\mathcal{D}$ into $K$ groups as $\{\mathcal{D}_1,\mathcal{D}_2, ..., \mathcal{D}_K\}$, 
with the consideration that the existing of single data point won't be able to make much difference on the behaviors of the target network $\mathcal{M}$. So it's more meaningful to build the cluster-level training data attribution under this circumstance.

Here the target network is initialized with parameters $\theta_0$ and subsequently trained on the dataset $\mathcal{D}$ for $\tau$ epochs, 
and the synset is for finetuning the trained target network for $T$ epochs,
resulting in updated parameters $\theta_\tau$ and $\Tilde \theta_{T}$.
The objective is formulated as:
\begin{equation}
\begin{split}
&\!\mathcal{A}(\mathcal{D})\!:\! \theta_\tau\! =\! \arg\min_{\theta}\! \mathcal{L} (\theta, \mathcal{D})\!=\! \arg\min_{\theta}\!\sum_{k}\! \mathcal{L} (\theta,\! \mathcal{D}_k),\\
&\!\mathcal{U}(\mathcal{D}_\kappa)\!=\!\mathcal{A}(\underset{k\ne \kappa}{\bigcup}\!\mathcal{D}_k):  \theta_{\tau}^\kappa = \arg\min_{\theta} \sum_{k\ne \kappa}\! \mathcal{L} (\theta, \mathcal{D}_k), \\
&\!\mathcal{F}(\mathcal{S}_\kappa): \Tilde\theta_{T}^\kappa = \arg\min_{\theta} \mathcal{L} (\theta, \mathcal{S}_\kappa),  \\
& s.t. \quad \mathcal{S}_\kappa \!=\!\underset{\mathcal{S}_\kappa, |\mathcal{S}_\kappa|\ll |\mathcal{D}_\kappa|}{\arg\min} \big[\mathcal{D}ist(\theta_{\tau}^\kappa, \Tilde \theta_{\tau}^\kappa) \big],
\end{split}
\label{eq:offline}
\end{equation}
where $\kappa = \{1,2,...,K\}$ and $\mathcal{L}(\cdot,\cdot)$ is the loss for training the target network $\mathcal{M}$. $\mathcal{A}$ stands for the learning process with $\tau$ epochs, $\mathcal{U}$ stands for the unlearning process (equivalent to training without the unlearn set with $\tau$ epochs), $\mathcal{F}$ stands for the fine-tuning process starting with $\Tilde\theta_{0} \leftarrow \theta_{\tau}$ with $T$ epochs.
The synset $\mathcal{S}=\{\mathcal{S}_1, \mathcal{S}_2, ..., \mathcal{S}_K\}$ is extremely small in scale comparing with the original dataset $\mathcal{D}$. To achieve this goal, our approach involves the minimization of the distribution distance, denoted as $\mathcal{D}ist(,)$, between the parameters of the synset fine-tuned model $\Tilde{\theta}_{\tau}^\kappa$, and the directly unlearned parameters $\theta_{\tau}^\kappa$.

Assuming that the target network parameters are updated through stochastic gradient descent for $t = 1,2,...,\tau$ epochs with a learning rate $\eta_a$, and the finetuning process of the target network spans $t = 1,2,...,T$ epochs with a learning rate $\eta_f$, we can reformulate the problem based on Eq.~\ref{eq:offline} as follows:
\begin{equation}
\begin{split}
\theta_{t+1}&\leftarrow \theta_t - {\eta_a} \nabla\mathcal{L}(\theta_t, \mathcal{D}),\\
\theta_{t+1}^\kappa&\leftarrow \theta_t^{\kappa} - {\eta_a }\sum_{k\neq \kappa}\nabla\mathcal{L}(\theta_t^{\kappa}, \mathcal{D}_k)\quad w.r.t. \quad \theta_0^\kappa =\theta_0,\\
\Tilde\theta_{t+1}^\kappa&\leftarrow \Tilde\theta_t^\kappa - {\eta_f }\nabla\mathcal{L}(\Tilde\theta_t^\kappa, \mathcal{S}_k)\quad w.r.t. \quad \Tilde \theta_0^\kappa =\theta_\tau,
\end{split}
\end{equation}
where $\nabla \mathcal{L}$ is the gradient computed on $\theta$. Based on it, we simplify the problem by setting $\eta= \eta_a = \eta_f$ and $\tau = T$.
In this way, we accumulate the gradients in the learning and finetuning process as:
\begin{equation}
\begin{split}
\theta_{\tau} &= \theta_0 - \eta\sum_t\nabla\mathcal{L}(\theta_t,\mathcal{D}),\\
\theta_{\tau}^\kappa &= \theta_0 - {\eta}\sum_t\sum_{k\neq \kappa}\nabla\mathcal{L}(\theta_t^\kappa,\mathcal{D}_k),\\
&= \theta_0 - {\eta}\sum_t \big[ \nabla\mathcal{L}(\theta_t,\mathcal{D})-\nabla\mathcal{L}(\theta_t^\kappa,\mathcal{D}_\kappa)\big],\\
\Tilde \theta_{\tau}^\kappa &= \theta_\tau - \eta\sum_t\nabla\mathcal{L}(\Tilde \theta_t^\kappa,\mathcal{S}_\kappa).
\end{split}
\label{eq:grad}
\end{equation}
\begin{figure}[t]
  \centering
  \includegraphics[width=0.4\textwidth]{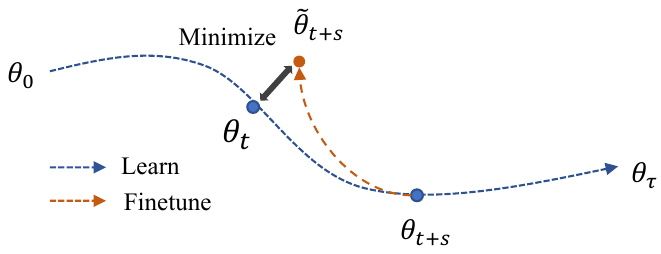}
  \caption{The proposed reverse gradient matching process. The synset is optimized by the reverse gradients.}
  \label{fig:unmatch}
\end{figure}

Note that our goal is to make $\Tilde \theta_{\tau}^\kappa\approx \theta_{\tau}^\kappa$, then Eq.~\ref{eq:grad} can be further simplified as:
\begin{equation}
\begin{split}
-\sum_t\nabla\mathcal{L}(\Tilde \theta_t^\kappa,\mathcal{S}_\kappa) = \sum_t\nabla\mathcal{L}(\theta_t^\kappa,\mathcal{D}_\kappa),\\
\end{split}
\label{eq:grad2}
\end{equation}
given that $\Tilde \theta_0^\kappa =\theta_\tau$ and $ \theta_0^\kappa =\theta_0$, the sufficient solution to Eq.~\ref{eq:grad2} is:
\begin{equation}
\begin{split}
   \nabla\mathcal{L}(\Tilde \theta_{\tau-t}^\kappa,\mathcal{S}_\kappa) &=- \nabla\mathcal{L}( \theta_{t}^\kappa,\mathcal{D}_\kappa),\\ 
\Rightarrow \quad \sum_{\kappa}\nabla\mathcal{L}(\Tilde \theta_{\tau-t}^\kappa,\mathcal{S}_\kappa) &= - \sum_{\kappa}\nabla\mathcal{L}( \theta_{t}^\kappa,\mathcal{D}_\kappa),\\
\Rightarrow \quad \sum_{\kappa}\nabla\mathcal{L}( \theta_{\tau-t},\mathcal{S}_\kappa) &= - \sum_{\kappa}\nabla\mathcal{L}( \theta_{t},\mathcal{D}_\kappa),
\end{split}
\end{equation}
where the synset in our proposed DDM is learnt for matching the reverse training trajectory while training the target network initialized from $\theta_0$ to $\theta_\tau$. This reverse gradient matching process is depicted in Fig.~\ref{fig:unmatch}.
Thus, synset $\mathcal{S}$ here is for predicting the parameters of the target network $\mathcal{M}$ that unlearns the $\kappa$-th data cluster $\mathcal{D}_\kappa$, which is achieved by directly finetuning the target network with the synset $\mathcal{S}$.

We constrain the scale of the synset to ensure efficient storage and fine-tuning process.
Motivated by the idea of dataset condensation~\cite{zhao2021dataset} with gradient matching, we propose the reverse gradient matching to distill and store the gradient information to a couple of synthetic images $\mathcal{S}$ ($|\mathcal{S}|\ll |\mathcal{D}|$). The synset $\mathcal{S}$ is optimized by:
\begin{equation}
\begin{split}
\!\underset{|\mathcal{S}|\!=\!K\times ipc}{\arg\min}\!\sum_{t}\!\sum_{\kappa}\mathcal{D}ist\big(\nabla\mathcal{L}( \theta_{\tau\!-\!t},\mathcal{S}_\kappa), \!-\!\nabla\mathcal{L}( \theta_{t},\mathcal{D}_\kappa) \big),\\
\end{split}
\label{eq:synset}
\end{equation}
where for each data cluster $\mathcal{D}_\kappa$, we learn a corresponding $\mathcal{S}_\kappa$ which contains $ipc$ images. In experiments, we set $ipc = 1$ and using the cosine distance for $\mathcal{D}ist(\cdot)$.

Why do we choose \textbf{reverse gradient matching over gradient matching}? There are two main reasons:
\begin{itemize}
\item \textbf{Enhanced matching performance.} Since the number of unlearn set is smaller than the whole set and the optimization of data  matches the initial stage of the learning trajectory, making the accumulated trajectory error much smaller using our proposed reverse gradient matching. Detailed evidence supporting this claim is provided in the supplementary materials.
\item \textbf{Improved Privacy Protection:} While traditional data distillation using gradient matching offers a degree of privacy protection for the dataset~\cite{dong2022privacy}, the distilled images still retain distinguishable patterns of the main object, posing privacy risks. In contrast, images synthesized using reverse gradient matching exhibit no explicit patterns, thus ensuring a higher level of privacy protection. Detailed comparisons in this regard are presented in the experimental results.
\end{itemize}

\subsection{Online Evaluation}
\label{sec:online}
During the online evaluation stage, both the synset $\mathcal{S}$ and the target network $\mathcal{M}$ are available, allowing for the evaluation of specific model behaviors.

The primary concept behind online evaluation is to employ leave-one-out cross-validation, which entails systematically perturbing the training dataset $\mathcal{D}$ by removing specific training samples. This process helps analyze the resulting impact on the model's performance.
Here we take studying the influence function for example, which is a typical task for analyzing the model's data sensitivity, offering insights into its robustness and decision boundaries. To be concrete, given a test sample $\{x_{t},y_t\}$, 
The corresponding prediction result as $\Tilde y_t$, where the target model is trained on the original whole dataset $\mathcal{D}$.
The objective here is to directly build the relationship with the network prediction $\Tilde y_t$ and the training data $\mathcal{D}$ (dataset$\rightarrow$target network$\rightarrow$prediction) by the attribution matrix $W$:
\begin{equation}
    \Tilde {y_t}_{(p_t\circ\mathcal{D})} = W\cdot p_t + b, \quad p_t \subseteq \{0,1\}^K,
\end{equation}
where $p_t$ stands for the perturbation operation over the dataset, and $p_t(\kappa)=0$ denotes the deletion of that data cluster $\mathcal{D}_\kappa$ from the training set. And $\Tilde {y_t}_{(P_t\circ\mathcal{D})}$ denotes the prediction by the target network, which is trained from scratch using the dataset $p_t\circ\mathcal{D}$.

Then, the attribution matrix $W$ calculated from perturbing the training data can be calculated as: 
\begin{equation}
\begin{split}
\underset{W}{\arg\min}\sum_{p_t\subseteq P_t} \beta_{p_t}\cdot \mathcal{D}ist\big( \Tilde {y_t}_{(p_t\circ\mathcal{D})}, W\cdot p_t\big ),
\end{split}
\label{eq:all}
\end{equation}
where ${p_t}$ is randomly sampled from the $\{0,1\}^K$,
$\beta_{p_t}$ represents the weights corresponding to the number of 0s in ${p_t}$.
The distance function $\mathcal{D}ist(\cdot)$ is set as the L2 norm distance for measuring influence function of the model. And the attribution matrix $W\subseteq \mathbb{R}^{K\times |y_t|}$, signifies the contribution of each training data cluster to the confidence scores of each label in the target network's prediction for the test sample $x_t$. 
Let $P_t$ denote the perturbation set. To calculate the attribution matrix $W$, a minimum of $K$ perturbations is required, such that $|P_t| \ge K$, applied to the training data.
 
The main difficulty in Eq.~\ref{eq:all} lies in obtaining $|P_t|$ new trained models training with the perturbed dataset $p_t\circ\mathcal{D}$, so as to get the corresponding inference $\Tilde {y_t}_{(p_t\circ\mathcal{D})}$. 
Recall that during the offline training process, we already got the distilled synthetic data $\mathcal{S}_\kappa$  for each data cluster $\mathcal{D}_\kappa$, which could fast unlearn $\mathcal{D}_\kappa$ from the target network.
And the model parameters with the perturbed dataset could be finetuned with the synset as:
\begin{equation}
    \theta_{p_t} \leftarrow \underset{\kappa\in \{1,2,...,K\}, p_t(\kappa) =0}{\mathcal{F}} (\mathcal{S}_\kappa).
\end{equation}
As a result, in the offline evaluation stage, we solve this difficulty by eliminating each cluster of training data from the target network, which is further accelerated by our proposed  reverse gradient matching.

\textbf{Accelerate with hierarchical distilled datamodel.}
To expedite the online evaluation process, we implement a hierarchical data distillation approach. This strategy encompasses the distillation of both the class-wise datamodel (with $K = |y|$) and the cluster-wise datamodel (with $K = |y| \times c$, where each label's data is partitioned into $c$ clusters).

By following this approach, we can construct both the class-wise and the cluster-wise attribution matrices. This approach accelerates the analysis of model behavior, including tasks such as identifying the most influential training data. This is achieved by initially pinpointing the class-wise data points and subsequently calculating the training matrix within each class.

\newcommand{\up}[1]{\scriptsize\textcolor{red}{#1}}

\begin{table*}[htp]
\centering 
\small
\caption{Ablation study on the influence analysis of certain test samples on MNIST, CIFAR10 and CIFAR100 datasets. We locate to the source data considering three distance functions. We report the value$\times100$ for $\mathcal{D}ist_1$  in the table, larger is better and tiny number in red donates the improvement or drop compared with `Random Select'.}
\begin{tabular}{p{25mm}<{\centering}p{12mm}<{\centering}p{12mm}<{\centering}p{12mm}<{\centering}p{12mm}<{\centering}p{12mm}<{\centering}p{12mm}<{\centering}p{12mm}<{\centering}p{12mm}<{\centering}p{12mm}<{\centering}}
\toprule 
\multirow{2}{*}{\textbf{Method}} & \multicolumn{3}{c}{\textbf{MNIST}} & \multicolumn{3}{c}{\textbf{CIFAR10}} & \multicolumn{3}{c}{\textbf{CIFAR100}} \\ \cmidrule(r){2-4}\cmidrule(r){5-7}\cmidrule(r){8-10}
& $\mathcal{D}ist_1$   & $\mathcal{D}ist_2$  & $\mathcal{D}ist_3$  & $\mathcal{D}ist_1$    & $\mathcal{D}ist_2$   & $\mathcal{D}ist_3$  & $\mathcal{D}ist_1$   & $\mathcal{D}ist_2$  & $\mathcal{D}ist_3$     \\ \midrule \midrule
Random Select & 3.3& 0.43& 0.07&2.7& 0.34&0.54&2.3&0.54 & 0.8 \\
Predict-based&5.5 \up{+2.2}& - & -& 3.1 \up{+0.4} &  - & -& 3.9 \up{+1.6}&  - & -\\
Clustering-based  & 6.2 \up{+2.9}& - & -& 2.5 \up{-0.2}&   - & - &3.6 \up{+1.3}& - & -\\
\midrule
DDM w/o cluster& 9.1 \up{+5.8}& 0.55 \up{+0.12}& 0.11 \up{+0.04}& 5.9 \up{+3.2} & 0.57 \up{+0.23}& 0.78 \up{+0.24}& 3.5 \up{+1.2}& 0.77 \up{+0.23}& 1.2 \up{+0.4}\\
DDM-match & \textbf{10.8 \up{+7.5}}  & \textbf{0.73 \up{+0.30}} & \textbf{0.13 \up{+0.06}} & 5.8 \up{+3.1}& 0.76 \up{+0.42}& 0.81 \up{+0.27} & 4.8 \up{+2.5} & 0.71 \up{+0.17}& \textbf{1.6 \up{+0.8}} \\
DDM-full (ours) & \textbf{10.8 \up{+7.5}} & \textbf{0.73\up{+0.30}} & \textbf{0.13 \up{+0.06}} & \textbf{6.8 \up{+4.1}} & \textbf{0.81 \up{+0.41}} & \textbf{0.92 \up{+0.38}}  & \textbf{5.3 \up{+3.0}} & \textbf{0.88 \up{+0.34}} & \textbf{1.6 \up{+0.8}} \\ \bottomrule
\end{tabular}
\label{tab:all}
\end{table*}

\begin{algorithm}[hb]
\caption{The Proposed DDM Framework}
\small
\label{alg::algo}
\vspace{-1em}
\begin{flushleft}
----------------------------\textit{\textbf{ Offline Training }}----------------------------\\
 \textbf{Input:} $\mathcal{D}$: training set; $\mathcal{M}$: target model; $\{\theta_0,\theta_1,...,\theta_{\tau}\}$: training trajectory of the target network ; 
 $s$: trajectory step.
\end{flushleft}
\begin{algorithmic}[1] 
\State Divide the training data $\mathcal{D}$ into $K$ clusters;
\State Randomly initialize $K$ synthetic samples, formed as $\mathcal{S}$;
\For{ {each} distillation step}
\State Choose random start from target trajectory: $\theta_t$ ($0\!\le\! t\!< \!\tau)$;
\State Choose the end from target trajectory: $\theta_{t+s}$ ($t + s< \tau)$;
\For{ $\kappa = 1,2,...,K$ }
\State Calculate the gradients on real data: $\nabla\mathcal{L} (\theta_{t}; \mathcal{D}_{\kappa})$;
\State Calculate the gradients of the synset $\nabla\mathcal{L}( \theta_{t+s},\mathcal{S}_\kappa)$;
\State $\min \mathcal{D}ist(\nabla\mathcal{L} \big(\theta_{t}; \mathcal{D}_{\kappa}),-\nabla\mathcal{L}( \theta_{t+s},\mathcal{S}_\kappa)\big)$ to update $ \mathcal{S}_{\kappa}$;
\EndFor
\EndFor
\end{algorithmic}
\begin{flushleft}
\textbf{Output:} Cluster-wise synthetic data $\{\mathcal{S}_{1},\mathcal{S}_{2},...,\mathcal{S}_{K}\}$.
\end{flushleft}
\vspace{-1em}
\begin{flushleft}
--------------------------\textit{\textbf{ Online Evaluation }}--------------------------\\
\textbf{Input:} $\mathcal{M}$: target model; $\mathcal{S}$: synset; $x_t$: test sample.
\end{flushleft}
\begin{algorithmic}[1]
 \State Randomly sample perturbations $p_t$ to form $P_t$;
 \For{each $p_t$ \textbf{in} $P_t$}
 \State Perform perturbation $p_t$ on the synset $\mathcal{S}$ as $p_t\circ \mathcal{S}$; 
 \State Fine-tune $\mathcal{M}$ with $p_t\circ \mathcal{S}$;
 \State Input $x_t$ to the fine-tuned network and get $\Tilde y_t$;
\EndFor
\State Calculate the attribution matrix $W$ with Eq.~\ref{eq:all}.
\end{algorithmic}
\begin{flushleft}
\textbf{Output:} Attribution matrix $W$.
\end{flushleft}
\vspace{-1em}
\end{algorithm}

\subsection{Algorithm and Discussions.}

We depict the proposed algorithm including offline training and online evaluation in Alg.~\ref{alg::algo}.
During the offline training stage, we follow and modify the basic optimization framework of dataset distillation.  And we give the algorithm for evaluating the influence function with $x_t$ as input. 
 In the online evaluation phase, adjustments are made to accommodate different evaluation objectives. Importantly, the offline training process occurs only once and remains fixed for subsequent evaluations.

\section{Experiments}
\subsection{Experimental Settings}
\textbf{Datasets and networks.} 
We conduct our experiments on several standard image classification datasets: digit recognition on MNIST dataset~\cite{LeCun1998GradientbasedLA}, CIFAR-10 dataset, CIFAR-100 dataset~\cite{Krizhevsky2009LearningML} and TinyImageNet~\cite{Russakovsky2014ImageNetLS}.
Regarding the architectures of the target network, we evaluated various architectures, including AlexNetIN~\cite{Krizhevsky2012ImageNetCW}, ResNet18, ViT, and ConvNet.

\textbf{Training details and parameter settings.}
We implemented our experiments using the PyTorch framework.
In the default setting, unless otherwise specified, we set the number of clusters per class to $num_cluster = 10$. This implies that there are a total of $K = 100$ clusters for the MNIST and CIFAR-10 datasets, and $K = 1000$ clusters for the CIFAR-100 dataset.
For both class-wise and cluster-wise condensation, we used a single synthetic image per cluster. These synthetic images were initialized by randomly sampling real images, and standard data augmentation techniques were applied during training.
The learning rate for updating synthetic images was set to 10, while the learning rate for updating network parameters was set to 0.01. To perturb the training set $\mathcal{D}$, we set $|P_t| = K$.

\textbf{Evaluation metrics.}
To assess the attribution of training data to the behavior of the target network when influencing specific test samples, we investigate three influence objectives, each defined by a distinct distance metric. We randomly select 20 test samples ($|X_t|=20$) from the validation set of the training data and report the average distance metrics.
And in order to compare the accuracy of such built relationship, we use the exact-unlearn network for evaluation.
That is, after locating the data cluster $\mathcal{D}_{i}$ with the target influence objective,
we scratch train the unlearn network $\mathcal{M}^u$ on $\mathcal{D}/\mathcal{D}_{i}$, and get predictions as $y_t^u$.
We compute distance function $Avg\_dist$ regarding different types of influence analysis:
\begin{equation}
\begin{split}
       &Avg\_dist = \mathbb{E}_{x_t\sim X_t}\mathcal{D}ist_i\\
\!where\quad \mathcal{D}ist_1 &=\|y_t^u-\Tilde{y}_t\|^2,\mathcal{D}ist_2 = \ell_{ce}(y_t^u,{y}_t) , \\
  \mathcal{D}ist_3 &=1/ (1+\|y_t^u-\Tilde{y}_t\|^2)  , 
\end{split}
\label{eq:avgdist}
\end{equation}
where $y_t$ is the groundtruth label for $x_t$ and $\Tilde y_t$ is the output from the target network $\mathcal{M}$. For all three distance metrics, namely $\mathcal{D}ist_1$, $\mathcal{D}ist_2$, and $\mathcal{D}ist_3$, larger values are indicative of more significant influence. Specifically, $\mathcal{D}ist_1$ is designed to trace back to the training data points that have the most influence on the current prediction, $\mathcal{D}ist_2$ focuses on identifying those with the most influence on whether the model makes correct predictions using the cross entropy loss $\ell_{ce}$, and $\mathcal{D}ist_3$ is utilized to pinpoint the training data points with the least influence on the current prediction.

For evaluation objectives other than the influence function of specific test samples, relevant metrics are provided within the corresponding experimental analysis part.

\subsection{Experimental Results}

\textbf{DDM could be used for training data influence analysis.}
By telling us the training points ``responsible'' for a given prediction, influence functions reveal insights about how models rely on and extrapolate from the training data. 
For three different distance functions ($\mathcal{D}ist_1$, $\mathcal{D}ist_2$ and $\mathcal{D}ist_3$), we calculate different weight matrix from Eq.~\ref{eq:all} by replacing the corresponding distance function, obtaining $W^1,W^2,W^3$. And then we locate the corresponding $\mathcal{D}_i$, where $i = \arg\max_i W_i$.
The ablation study regarding the three types of influence functions is depicted in Table~\ref{tab:all}, where `Random Select' denotes that we randomly choose $D_i$ from K clusters; `Predict-based' denotes we choose the datapoint $\mathcal{D}_i$ with the highest prediction similarity.
`Cluster-Based' denotes locating $\mathcal{D}_{i}$ by using the clustering strategy as we pre-process the dataset $\mathcal{D}$, which denotes the highest visual similarity;
`DDM w/o cluster' clusters $\mathcal{D}$ in $K$ clusters by sequence number not by k-means,
`DDM-match' denotes the DDM framework that uses the gradient matching loss during the offline training stage.
From the table, we observe that:
\begin{itemize}
    \item Methods categorized as `Predicted-based' and `Clustering-based' can serve as alternatives for evaluating $\mathcal{D}ist_1$ type inferences. While they turn to be less accurate than our DDM. It further proves that visual similarity between two images does not fully capture the influence of one on the other in terms of model behavior~\cite{ilyas2022datamodels}. 
    In addition, they are limited in their ability to provide a comprehensive analysis of $\mathcal{D}ist_2$ and $\mathcal{D}ist_3$. This highlights the potential of our proposed DDL model to evaluate a wider range of model behaviors.
    \item Comparing the results with `DDM w/o cluster' and `DDM-full', it becomes evident that the clustering strategy offers a more accurate method for pinpointing influential training data.
    \item Upon comparing the results with `DDM-match' and `DDM-full', it is apparent that optimizing the synset with gradient matching produces favorable outcomes on simpler datasets like MNIST. However, as the dataset complexity increases, this approach exhibits a decline in performance, falling behind the optimization with reverse gradient matching.
\end{itemize}

\begin{table}[t]
\small
\centering
\caption{Comparative experimental results with other works on MNIST, CIFAR10 and CIFAR100 datasets, regarding $\mathcal{D}ist_1$ influence. }
\begin{tabular}{p{24mm}<{\centering}ccc}
\toprule
\textbf{Method} & \textbf{MNIST}&\textbf{CIFAR-10} &\textbf{CIFAR-100}\\ \midrule \midrule
Random Select      &  3.3 & 2.7 & 2.3    \\
Koh et al.~\cite{koh2017understanding}& 10.0 & 5.6 &2.6 \\
FASTIF ~\cite{guo2020fastif}& 9.8 & 6.5&2.4 \\
Scaleup~\cite{schioppa2022scaling} & 10.4 &6.5&3.9 \\
DDM & \textbf{10.8} & \textbf{6.8} & \textbf{5.3} \\\bottomrule       
\end{tabular}
\label{tab:sota}
\end{table}
\begin{table}[t]
\small
\centering
\caption{Detecting useless training data for the target network. }
\begin{tabular}{p{24mm}<{\centering}ccccc}
\toprule
\textbf{Percentage} & \textbf{0\%}&\textbf{1\%} &\textbf{10\%}&\textbf{20\%}&\textbf{50\%}\\ \midrule \midrule
Random Select   & 95.7 & 95.8 &92.8  & 74.6& 65.9 \\
Koh et al. & 95.7 & 95.9 & 93.7& 81.5 & 74.3  \\
FASTIF  & 95.7 & 95.5 & 94.1& 79.6&74.0\\
Scaleup & 95.7&  96.0& 95.2&82.9 &73.3 \\
DDM & 95.7 & 96.2 & 95.9 & 85.4 & 79.3 \\\bottomrule       
\end{tabular}
\label{tab:percent}
\vspace{-1em}
\end{table}

\begin{figure}[t]
\centering
\includegraphics[scale = 0.39]{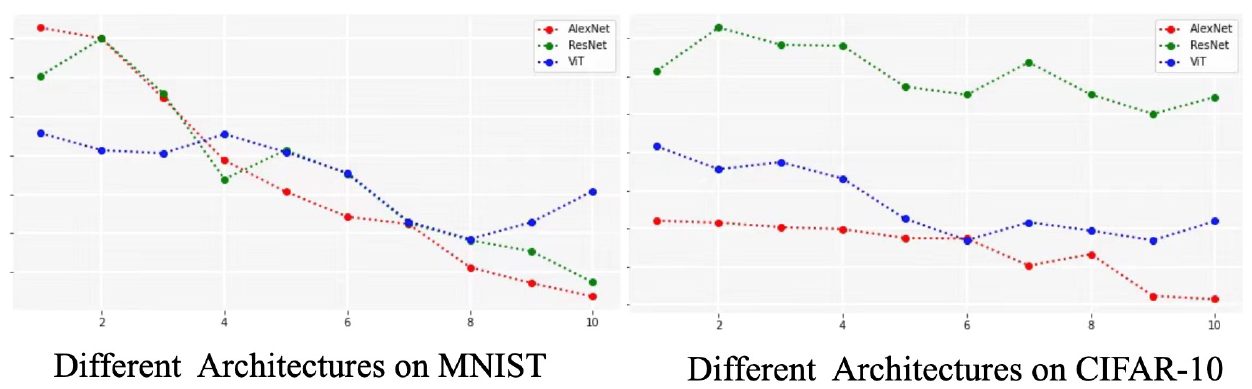}
\caption{Comparison of the training data attribution weights calculated form different network architectures. In the figure, we show the class-wise weights.}
\label{fig:diffarch}
\vspace{-2em}
\end{figure}

\textbf{How do different target network architectures affect DDM?}
We have conducted our proposed DDM framework into several different network architectures, including AlexNetIN, ResNet18\_AP~\cite{he2016deep} and simple ViT~\cite{dosovitskiy2020image}. 
We calculate the training data attribution weights for each class of training data for measuring the model behaviors on $\mathcal{D}ist_1$. The test input is a batch of images with the groundtruth label of `2'.
The experimental results are conducted on the MNIST dataset and the CIFAR-10 dataset, which are depicted in Fig.~\ref{fig:diffarch}.
From the figure, observations can be drawn that:
\begin{itemize}
    \item All the networks with different architectures trace to the similar source training data (with the highest value with label `2');
    \item For the simple classification task in MNIST, the training data attribution matrices look similar among all the architectures;
    \item For a more difficult task in CIFAR10, the training data attribution matrices look also similar in the trend, but vary in the absolute weight values among all the architectures.
\end{itemize}

\textbf{Comparing DDM performance with other works.} 
The comparative results with existing works are presented in Table~\ref{tab:sota}. We compare the $\mathcal{D}ist_1$ influence with three other works. To evaluate, we identify the most influential data point and calculate the average distance metrics.
It is evident that our method demonstrates greater accuracy in locating these influential data points.

\begin{table}[t]
\small
\centering
\caption{The new trained network's accuracies comparison. We compare the networks fine-tuning with the proposed DDM and gradient matching synthetic images. }
\begin{tabular}{p{23mm}<{\centering}ccc}
\toprule
\textbf{Dataset}  & \textbf{Method}    & \multicolumn{1}{c}{\textbf{Acc. ($\theta_0$)}}     & \multicolumn{1}{c}{\textbf{Acc. ($\theta_\tau$)}}     \\ \midrule \midrule
\multirow{3}{*}{MNIST}                            & Normal    & \multicolumn{1}{c}{12.5} & \multicolumn{1}{c}{95.7} \\
                                                  & DDM-Match & \multicolumn{1}{c}{12.6} & \multicolumn{1}{c}{85.0} \\
                                                  & DDM       & \multicolumn{1}{c}{0.6}  & \multicolumn{1}{c}{95.7} \\\midrule
\multirow{3}{*}{CIFAR 10}                         & Normal    & \multicolumn{1}{c}{12.6} & \multicolumn{1}{c}{85.0} \\
                                                  & DDM-Match & 12.6                     & 40.2                     \\
                                                  & DDM       & 15.5                     & 85.0                     \\\midrule
\multirow{3}{*}{CIFAR100}     & Normal    & 2.2                      & 56.1                     \\
\multicolumn{1}{l}{}                              & DDM-Match & 2.2                      & 23.5                     \\
\multicolumn{1}{l}{}                              & DDM       & 0.8                      & 56.1                     \\\midrule
\multirow{3}{*}{TinyImageNet} & Normal    & 1.4                      & 37.5                     \\
\multicolumn{1}{l}{}                              & DDM-Match & 1.4                      & 7.8                      \\
\multicolumn{1}{l}{}                              & DDM       & 0.1                      & 37.5  \\\bottomrule                  
\end{tabular}
\label{tab:mandun}
\end{table}
\textbf{DDM could be used as model diagnostic for low-quality training samples.}
In addition to analyzing the influence functions for specific test samples, the proposed DDM also offers a comprehensive model of the overall performance of the target network.
We randomly sample $10\%$ samples and calculate the 
The experimental results are depicted in Table~\ref{tab:percent}, which is conducted on MNIST dataset.
As depicted in the figure, the deletion of $10\%$ of the training data actually improves the network's final performance. Therefore, our proposed DDM framework succeeds in model diagnostics by detecting and removing low-quality training samples.

\textbf{DDM meets the privacy protection demand.}
\begin{figure*}[t]
\centering
\includegraphics[scale = 0.6]{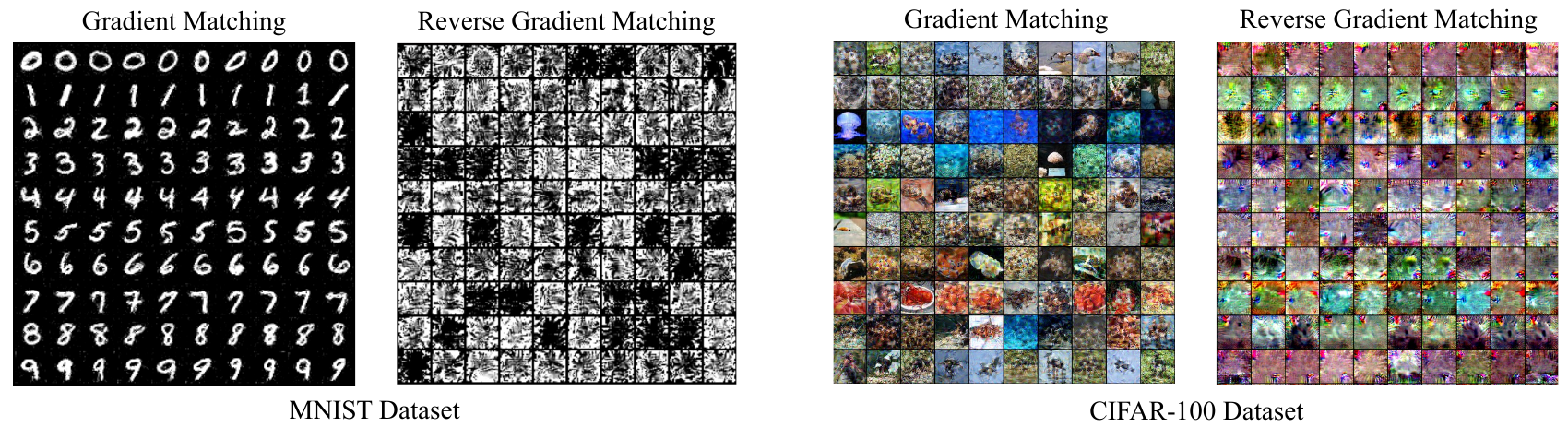}
\vspace{-1em}
\caption{Visualization of condensed 10 image/class with ConvNet for MNIST (a) and CIFAR-100 (b). We compare the visualization results between gradient matching and reverse gradient matching. Each column represents a condensation of a cluster.}
\label{fig:vis}
\vspace{-1em}
\end{figure*}
To substantiate our previous assertion that the proposed reverse gradient matching enhances privacy protection, we conducted a comparison between the distilled samples generated using traditional gradient matching and our novel reverse gradient matching, as illustrated in Fig.~\ref{fig:vis}.
As evident from the visualization, in gradient matching data distillation, the synthetic images retain the characteristic features of the training set images, thus potentially revealing training data through these conspicuous patterns, particularly noticeable in the MNIST dataset.
In contrast, in the visualization results of our reverse gradient matching, the distinctive features of the images are replaced by several indistinct patterns, akin to a form of obfuscation. This implies that, especially in scenarios with privacy concerns, our proposed DDM framework can be safely employed by directly releasing the synset, providing enhanced privacy protection for the original training data.

\textbf{DDM could be used as a quick unlearn method.}
We also assert that the proposed reverse gradient matching improves matching performance, which is experimentally verified in Table~\ref{tab:mandun}.
It's worth noting that traditional gradient matching begins by matching the initial state of the target network, resulting in the same `Acc.($\theta_0$)' as normal training. However, for more complex datasets (e.g., TinyImageNet), it struggles to match the final performance of the target network, `Acc. ($\theta_\tau$)'. In contrast, our proposed DDM commences from the final state of the target network and also effectively matches the initial performance of the target network. Thus, we contend that the proposed DDM significantly enhances matching performance.
\section{Conclusion}
In this paper, we introduce a novel framework known as DDM that facilitates a comprehensive analysis of training data's impact on a target machine learning model.
The DDM framework comprises two key stages: the offline training stage and the online evaluation stage. During the offline training stage, we propose a novel technique, reverse gradient matching, to distill the influence of training data into a compact synset. In the online evaluation stage, we perturb the synset, enabling the rapid elimination of specific training clusters from the target network. This process culminates in the derivation of an attribution matrix tailored to the evaluation objectives.
Overall, our DDM framework serves as a potent tool for unraveling the behavior of machine learning models, thus enhancing their performance and reliability. 

Future research could extend the application of the DDM framework to diverse machine learning tasks and datasets. These applications could encompass fields as varied as natural language processing, reinforcement learning, computer vision, and beyond. The versatility of DDM offers opportunities to gain deeper insights into model behaviors, data quality, and training dynamics in these domains.
\section*{Acknowledgement}
This project is supported by the Ministry of Education Singapore, under its Academic Research Fund Tier 2 (Award Number: MOE-T2EP20122-0006),
and the National Research Foundation, Singapore, under its AI Singapore Programme (AISG Award No: AISG2-RP-2021-023).
{\small
\bibliographystyle{ieeenat_fullname}

}

\clearpage
\setcounter{page}{1}
\maketitlesupplementary

In this document, we present supplementary materials that couldn't be accommodated within the main manuscript due to page limitations. Specifically, we offer additional details on the proposed DDM framework, elucidating the computation of the hierarchical attribution matrix through our framework and the methodology employed for data clustering.

\section{More Details of DDM Framework}

\subsection{Data Clustering of DDM}
To enhance tracing performance, we concentrate on calculating the weighting matrix for each batch of data, rather than for each individual image. 
That is, the total cluster number $K$ is set as $L<K<|\mathcal{D}|$ ($L$ is the total number of the class).
This approach is taken as the impact of a single image becomes negligible when training with a large number of images.

In our proposed DDM framework, we employ data clustering to partition the training data into several distinct batches. Specifically, we utilize a pre-trained feature extractor to embed the training images into the same feature space. Subsequently, we apply K-means clustering to the features, denoting each cluster of data as $\mathcal{D}_{c,k}$, where $l = 1,2,...,L$ and $c = 1,2,...,C$. This implies that we cluster the images within a class into $C$ clusters.
For each class of data, we set the number of clusters $C$ to be 10.

\subsection{Accelerating with Hierarchical DDM}
\label{sec:hierarchical}
As stated in the main paper, to enhance the efficiency of online evaluation, we employ hierarchical attribution calculation, which expedites the leave-one-out retraining process. 
Note that on the offline training stage, both the class-wise and cluster-wise synset are learned, where the class-wise one is denoted as $\mathcal{S}_{class}=\{\mathcal{S}_1,\mathcal{S}_2,...,\mathcal{S}_L\}$ and the cluster-wise one is denoted as  $\mathcal{S}_{cluster}=\{\mathcal{S}_{l,1},\mathcal{S}_{l,2},...,\mathcal{S}_{l,C}\}_{l=1}^L$ and $K = L \times C$.
With this hierarchical synset, we don't need to apply the retraining to the entire perturbation set $|P_t|$. Instead, it is calculated as:
\begin{equation}
\begin{split}
l \leftarrow \underset{l}{\arg\max} W_l,\quad 1 \le l \le L,\\
c \leftarrow \underset{l}{\arg\max} W_{l,c},\quad 1\le c\le C
\end{split}
\label{eq:trace}
\end{equation}
In the first equation, we require $|P_t| = L$, and in the second one, we need $|P_t| = C$. Therefore, with this hierarchical synset, we only need to retrain the networks $C+L$ times, significantly reducing the original $C \times L$ retraining to $C+L$.

In the standard setting for the CIFAR-10 dataset, where $K=10\times 10$, and for the CIFAR-100 dataset, where $K=100\times 10$, the hierarchical synset accelerates the process by 5 times in the CIFAR-10 dataset and 10 times in the CIFAR-100 dataset.

\subsection{Reverse Gradient Matching for Matching Performance Enhancement}
In the main paper, we claim that our proposed reverse gradient matching provides enhanced matching performance, which has been proved in the experiment (`DDM could be used as a quick unlearn method').
The reasons for it can be analyzed by the accumulated errors. 
Since we performance the leave-one-out retraining, which means that each time we unlearn one data cluster. Thus, for unlearning the data cluster $\mathcal{D}_{\kappa}$, the errors of our proposed reverse gradient matching can be denoted as:
\begin{equation}
    \epsilon_{\kappa} = \sum_t \big|\nabla \mathcal{L}(\theta_{\tau-t},\mathcal{S}_{\kappa}) + \nabla \mathcal{L}(\theta_{t},\mathcal{D}_{\kappa})\big|,
\label{eq:error}
\end{equation}
where $1<\kappa < K$ and $\mathcal{S}_{\kappa}$ is corresponding synset that matches the data cluster $\mathcal{D}_{\kappa}$ with reverse gradients. And denote $\mathcal{X}_{\kappa}$ as the corresponding synset that matches the data cluster $\mathcal{D}_{\kappa}$ with normal gradients, then the accumulated errors could be denotes as:
\begin{equation}
    \epsilon_{\kappa} = \sum_t \sum_{k\ne \kappa} \big|\nabla \mathcal{L}(\theta_{t},\mathcal{X}_{k}) - \nabla \mathcal{L}(\theta_{t},\mathcal{D}_{\kappa})\big|,
\end{equation}
which, compared with Eq.~\ref{eq:error}, introduces an additional summation operation, resulting in larger accumulated error values.

\begin{table*}[t]
\centering
\small
\caption{The detailed settings in the experimental implementation.}
\begin{tabular}{cccccccc}
\toprule
\multirow{2}{*}{\textbf{Dataset}} & \multirow{2}{*}{\textbf{Network}} & \multicolumn{6}{c}{\textbf{Settings}}                               \\ \cmidrule{3-8}
 &  & lr\_net & lr\_img & Training Epochs &Step for Synset & Step Length & Dist \\ \midrule
\multirow{2}{*}{MNIST} &  ConvNet/AlexNet/ResNet   &  0.001   &  0.1  &    30     &50     &   4   & Cosine Dist   \\
  &Simple ViT   &   0.001   &  0.005  &      30         &   50& 4    & MSE        \\\midrule
\multirow{2}{*}{CIFAR10} &  ConvNet/AlexNet/ResNet   &  0.01 & 0.1  & 30 &   50  & 4  & Cosine Dist   \\
  &Simple ViT   &  0.01   &  0.005  &      30   &   50    &   4   & MSE       \\ \midrule
  \multirow{2}{*}{CIFAR100} &  ConvNet/AlexNet/ResNet   &   0.01 & 0.1  & 30  &   50  & 4   & Cosine Dist  \\
  &Simple ViT   &  0.01   &  0.005  &      30   &   50    &   4 & MSE         \\\midrule
  \multirow{2}{*}{TinyImgNet} &  ConvNet/AlexNet/ResNet   &  0.01 & 0.1  & 30  &   50  & 4  & Cosine Dist   \\
  &Simple ViT   &  0.01   &  0.005  &      30   &   50        &   4  & MSE        \\\bottomrule     
\end{tabular}
\label{tab:set}
\end{table*}

\section{Experimental Setting}

The experimental setting of the proposed DDM framework is depicted in Table~\ref{tab:set}, where the dataset information, network backbones (ConvNet, AlexNet, ResNet and Simple Vit) are listed. Two different distance functions $\mathcal{D}ist(\cdot)$ are utilized in Eq.~\ref{eq:synset} of the main paper.

\begin{figure*}[t]
\centering
\includegraphics[scale = 0.57]{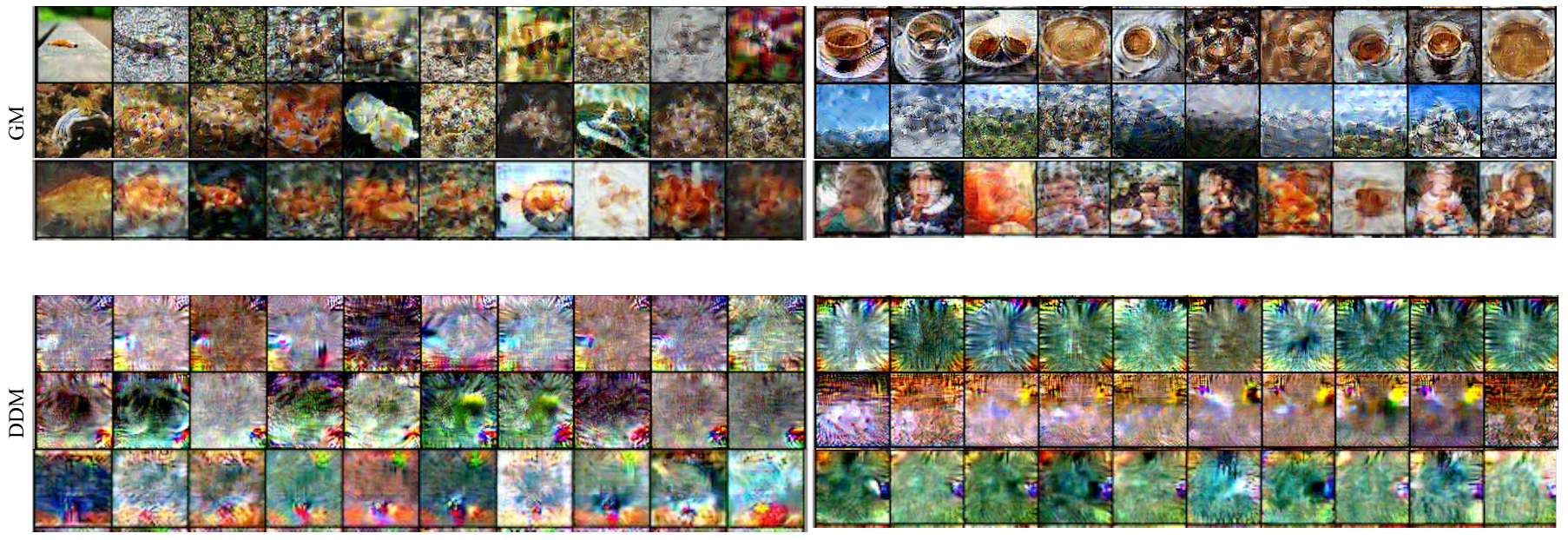}
\caption{Visualization of condensed 10 image/class with ConvNet for TinyImageNet dataset. We compare the visualization
results between gradient matching (GM) and reverse gradient matching (DDM). In each visualization, each column represents a condensation of a
cluster.}
\label{fig:visimage}
\end{figure*}

\section{More Experiments}

\begin{figure}[t]
\centering
\includegraphics[scale = 0.58]{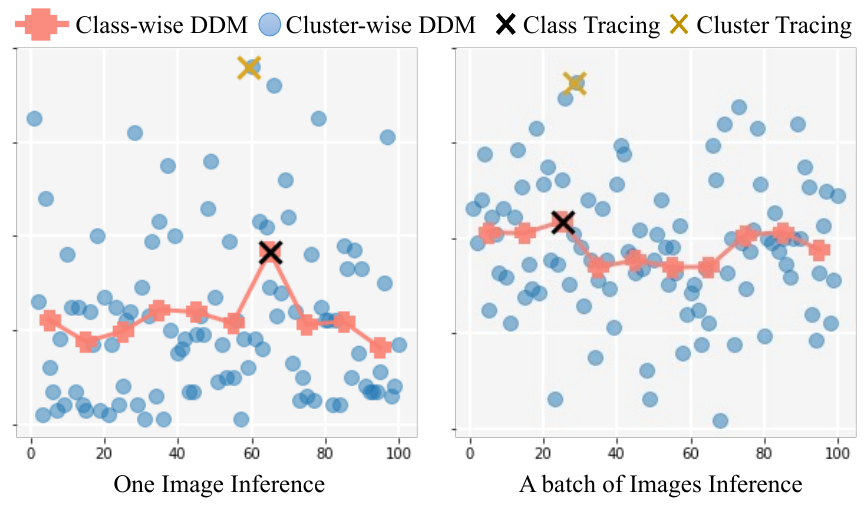}
\caption{Locating the most influential data points, where the results are computed with distance function $Dist_1$ on CIFAR10 dataset. We firstly locate the class with the class-wise synset and then to the cluster with the cluster-wise synset of that class.}
\label{fig:cluster}
\end{figure}
\begin{figure}[t]
\centering
\includegraphics[scale = 0.51]{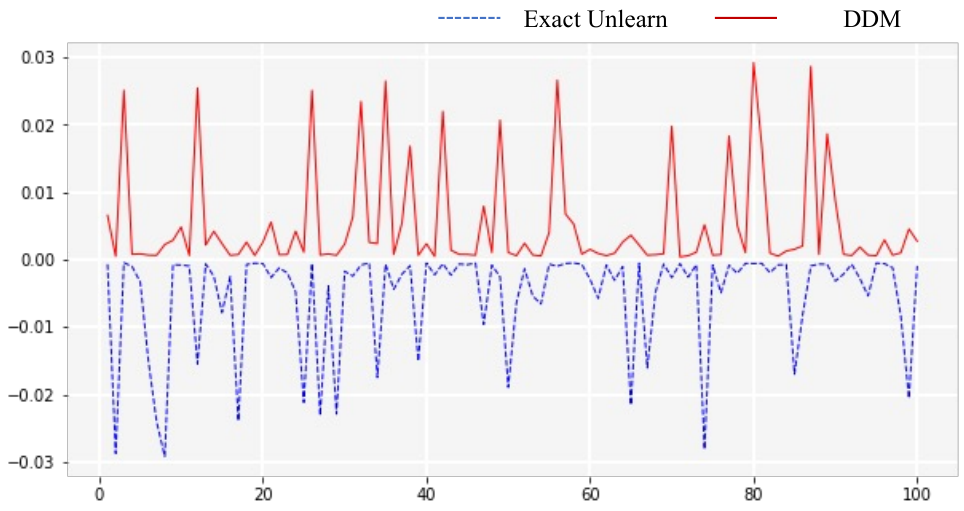}
\caption{Comparing the empirical error between the exact unlearn model and the DDM model. We scale and flip `exact unlearn'  for better visual comparison.}
\label{fig:erm}
\end{figure}

\begin{figure*}[t]
\centering
\includegraphics[scale = 0.52]{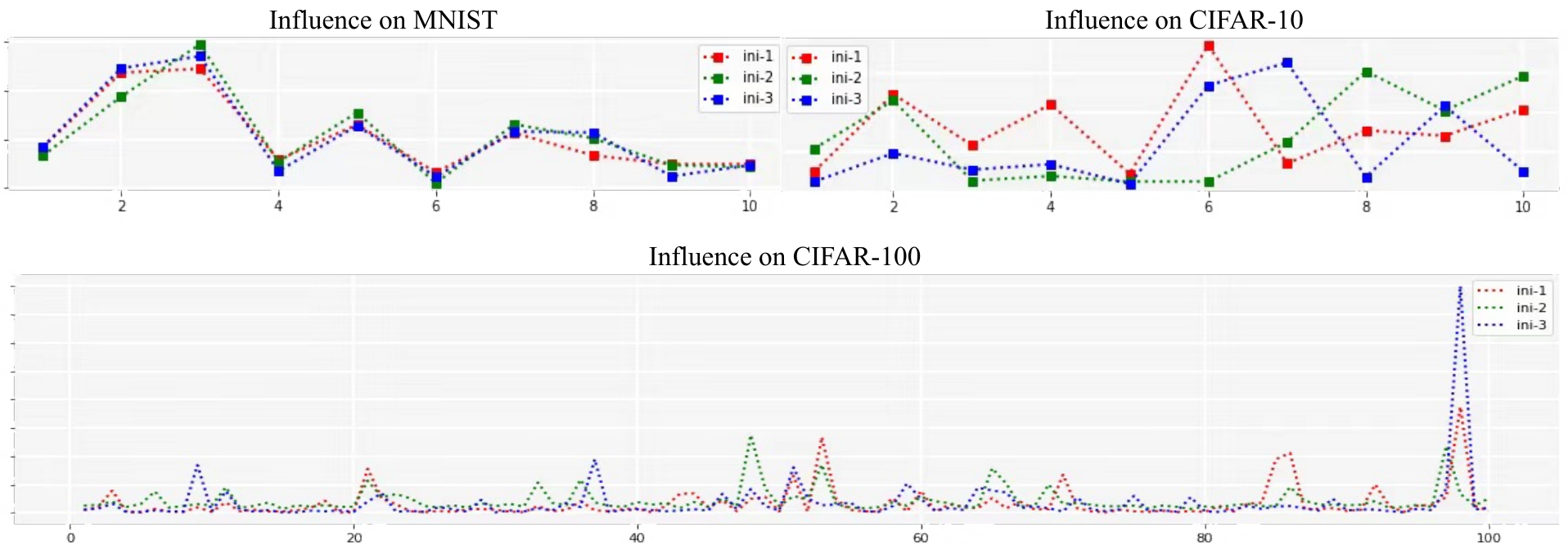}
\caption{Comparison of the training data attribution matrix with different initializations. The
experiments are conducted on MNIST, CIFAR-10 and CIFAR-100 datasets, with the ConvNet as the
base network.}
\label{fig:ini1}
\end{figure*}

\subsection{Class-wise DDM vs Cluster-wise DDM}

In our proposed DDM framework, both the class-wise and the cluster-wise synthetic images are obtained for efficient tracing by hierarchical search (details in Sec.~\ref{sec:hierarchical}).
In Fig.~\ref{fig:cluster}, we depict the tracing results (got by Eq.~\ref{eq:trace}) with the most influential datapoints in terms of $\mathcal{D}ist_1$, with $\|X_t\|=1$ (`one image inference' with groundtruth label as `5') and $\|X_t\|=256$ (`a batch of images inference' with all groundtruth labels as `2'). 
As can be observed, both the class-wise tracing and cluster-wise tracing give consistent tracing results.
In addition, DDM works on both the single image inference (gt label belongs to the 5-th label) and multi-image inference cases (all gt labels belong to the 2-th label).
indicating the robustness of the DDM framework.

\subsection{More Visualization Results}
In Fig.~\ref{fig:vis} of the main paper, we visualize the synset on MNIST dataset and CIFAR-100 dataset. Here, we provide additional visualization results on the TinyImageNet dataset, as depicted in Fig.~\ref{fig:visimage}.

The figure provides additional evidence supporting the privacy protection capabilities of our proposed DDM, as the synsets exhibit no recognizable objects. Moreover, the uniqueness of synthetic images learned from each data cluster highlights the importance of the clustering operation.

\subsection{Comparing DDM for Machine Unlearning with Exact Unlearn}
As discussed in the method section, we optimize the synthetic images using the proposed reverse gradient matching, which are then employed to fine-tune the target network to mitigate the impact of specific data clusters.
It's important to note that we don't anticipate the DDM framework to perfectly mimic the exact unlearned model. Instead, our focus is on whether it can capture important characteristics of the model.
In this experiment, we use empirical error for comparison, as depicted in Fig.~\ref{fig:erm}. In the figure, we scale and flip `exact unlearn', which, in an ideal situation, should be symmetric to `DDM'. As observed, they are roughly symmetrical, indicating that the DDM models can serve as a surrogate for analyzing the target model and perform well in eliminating certain data points compared to the exact unlearn.

\subsection{How Did Different Initializations Influence the Network?}

Here, we investigate the impact of different initializations on the computed training data matrix. We perform this comparative experiment on the MNIST, CIFAR-10, and CIFAR-100 datasets trained using the ConvNet architecture. We consider three different initializations: `Kaiming' (ini-1), `Normal' (ini-2), and `xavier' (ini-3).

The experimental results are illustrated in Fig.~\ref{fig:ini1}, revealing the following observations:
\begin{itemize}
\item The training data attribution is robust across different initialization methods, yielding similar attribution matrices. This observation holds true for the MNIST, CIFAR-10, and CIFAR-100 datasets.
\item With an increase in the size of the training data, the attribution matrices learned from the three different initializations become more diverse. This divergence may arise from the selection of the basic ConvNet as the backbone, potentially leading to local optima.
\end{itemize}

\section{DDM for Analysis of Other Model Behaviors}
In the main paper, we detailed instructions on analyzing networks by identifying the most influential training data for certain test images. We emphasize that both the local and global behaviors of the network can be captured by the training data matrix when using specific distance functions defined by the certain evaluation objective.

Here, we present various distance functions for different evaluation objectives.

\begin{table}[t]
\small
\centering
\caption{Detecting noisy training data for the target network. We add random noise with perturbation norm of $0.1$ to each image.}
\begin{tabular}{p{24mm}<{\centering}ccccc}
\toprule
\textbf{Percentage} & \textbf{0\%}&\textbf{10\%} &\textbf{20\%}&\textbf{30\%}&\textbf{40\%}\\ \midrule \midrule
Random Select   & 78.4 & 70.5 & 68.8 & 61.3& 55.7 \\
DDM & 78.4 & 82.0 & 79.3 & 73.6 & 70.1 \\\bottomrule       
\end{tabular}
\label{tab:noise}
\end{table}

\subsection{Inference Function of Certain Test Samples}
The distance function is defined in Eq.~\ref{eq:avgdist} in the main paper. This function aims to identify which part of the training data is responsible for the final prediction. We provide different evaluation objectives, including those influencing the current predictions most/least and those contributing the most to making correct predictions. 

And the corresponding experiments have been listed in the main paper in Table~\ref{tab:all}.

\subsection{Model Diagnostic for Low-quality Training Samples}
 To identify essential part of data that contributes to the overall prediction ability. We have already displayed the corresponding experiments in Table~\ref{tab:percent} by sorting the training cluster with its influence to the final network performance.
 To be specific, we randomly choose part of training data as the validation set $\mathcal{V}$, then we could determine distance function for the model global performance evaluation as:
 \begin{equation}
    \sum_{x_t\in \mathcal{V}} \mathcal{D}ist_2 (x_t),
    \label{eq:dist2}
 \end{equation}
where $\mathcal{D}ist_2$ is pre-defined in Eq.~\ref{eq:avgdist} for identifying those with the most influence on whether the model makes correct predictions.

In addition to the previous experiment focused on removing locally low-quality data from the training set, we conducted an additional experiment involving the introduction of random noise to $20\%$ of the training data. Subsequently, we removed the data based on the evaluation objectives mentioned earlier.

The experimental results are presented in Table~\ref{tab:noise}. The table reveals that the addition of random noise has a detrimental effect on the performance of the target model. However, when selectively removing data based on our proposed DDM, the network's performance improves, demonstrating a more significant improvement compared to the `Random Select' approach.

 \subsection{Transferability Between Different Networks.}
 And we also find that our proposed DDM also provides to measure and improve the transferability between different networks.
 To be concrete, in the typical work like knowledge distillation, the student network can be trained by:
 \begin{equation}
 \mathcal{L}_{\text{total}} = (1 - \alpha) \cdot \mathcal{L}_{ce}(\mathcal{N}(x), y^t) + \alpha \cdot \mathcal{KL}(\mathcal{N}(x), \mathcal{M}(x)),
 \end{equation}
 where $\mathcal{N}$ represents the student network to be trained, $\mathcal{M}$ is the target network, $\mathcal{KL}$ denotes the KL-divergence, and $\alpha$ is the balancing weight. It is important to note that not all networks may experience performance improvement through such distillation, as there could be conflicts that hinder the overall performance enhancement.

\begin{table}[t]
\small
\centering
\caption{DDM for network tranferability while distillation.}
\begin{tabular}{p{24mm}<{\centering}ccccc}
\toprule
\textbf{Percentage} & \textbf{0\%}&\textbf{10\%} &\textbf{20\%}&\textbf{30\%}&\textbf{40\%}\\ \midrule \midrule
Random Select   & 94.1 & 93.5 & 88.7 & 87.3& 85.8 \\
DDM & 94.1 & 94.2 & 94.2 & 92.9 & 87.3 \\\bottomrule       
\end{tabular}
\label{tab:distill}
\end{table}

 We use the KL-divergence as the distance function and calculate the whole evaluation distance in the similar way as in Eq.~\ref{eq:dist2}.
The experiments are conducted on Table~\ref{tab:distill}.
In the table, we delete some percentages of data from the network training and distill it to train the student network. The experiments are conducted on CIFAR-10 dataset on ResNet-18, and the teacher network is optimized by knowledge undistillation. Thus, directly distill from the teacher would cause accuracy drop.
And deleting some samples from the training data could attack such knowledge distillation and improve the network performance.

 \subsection{To be Explored.} 
 
 In this paper, we distilled the training gradients to several synthetic images, which enables the fast impact elimination. Thus, it is possible to build the training data attributes with the distance function defined between networks, which shows great potential to explore other kinds of network behaviors.

\end{document}